\documentclass[letterpaper]{article} 
\usepackage{aaai2027}  
\usepackage[hyphens]{url}  
\usepackage{graphicx} 
\usepackage{natbib}  
\usepackage{caption} 
\usepackage{algorithm}
\usepackage{algorithmic}
\usepackage{multirow}
\usepackage{amsmath}
\usepackage{amssymb,amsfonts}
\usepackage{multirow,booktabs}
\usepackage{newfloat}
\usepackage{listings}
\DeclareCaptionStyle{ruled}{labelfont=normalfont,labelsep=colon,strut=off} 
\floatstyle{ruled}
\newfloat{listing}{tb}{lst}{}
\floatname{listing}{Listing}

\usepackage{booktabs}

\title{PCSDiff: Diffusion-Based Bias Correction and Super Resolution Toward Practical Operational Medium-Term Precipitation Forecast}
\author {
    Yuze Sun\textsuperscript{\rm 1,2}\equalcontrib,
    Shiyi Wang\textsuperscript{\rm 1}\equalcontrib,
    Jiancheng Pan\textsuperscript{\rm 1},
    Die Wang\textsuperscript{\rm 3},
    Andreas F. Prein\textsuperscript{\rm 3},
    Wentao Luo\textsuperscript{\rm 2},
    Linhan Jiang\textsuperscript{\rm 4},
    Jie Wu\textsuperscript{\rm 5}\corresponding,
    Quan Zhang\textsuperscript{\rm 6}\corresponding,
    Xiaomeng Huang\textsuperscript{\rm 1}\corresponding
    }

\affiliations{
    \textsuperscript{\rm 1}Department of Earth System Science, Ministry of Education Key Laboratory for Earth System Modelling, Institute for Global Change Studies, Tsinghua University, Beijing, China\\
    \textsuperscript{\rm 2}Huawei Technologies Co., Ltd\\
    \textsuperscript{\rm 3}Institute for Atmospheric and Climate Science, ETH Zürich, 8092 Zürich, Switzerland\\
    \textsuperscript{\rm 4}School of Information and AI, Beijing Forestry University, Beijing, China\\
    \textsuperscript{\rm 5}State Key Laboratory of Climate System Prediction and Risk Management/China Meteorological Administration Key Laboratory for Climate Prediction Studies, National Climate Centre, China Meteorological Administration, Beijing, 100081\\
    \textsuperscript{\rm 6}National Institute of Natural Hazards, Ministry of Emergency Management of China, Beijing 100085, China.\\
    
    wujie@cma.gov.cn, quanzhang\_zq@163.com, hxm@tsinghua.edu.cn

}

\begin{document}

\maketitle

\begin{abstract}
Medium‑range precipitation forecasts are impaired by persistent systematic biases, lead‑time‑dependent error accumulation, and coarse spatial resolution, restricting their reliability for flood‑drought risk assessment. Existing AI correction techniques lack dedicated modeling for multi‑day dynamic bias evolution and proper meteorological constraints, often generating over‑smoothed rainfall structures, and cannot meet operational deployment demands. This work introduces PCSDiff, a cascaded task‑decoupled diffusion framework targeting 10‑day precipitation bias correction and downscaling. To jointly counteract temporal error drifts and reconstruct physically plausible local precipitation details, PCSDiff integrates the Precipitation Intensity‑aware Multi‑branch Decoder (PIMD) module for dynamic multi‑day error mitigation using synoptic‑temporal features, followed by a two‑phase conditional diffusion super‑resolution module to restore fine‑scale precipitation patterns. Evaluated against CMA‑CRA observations over China after global‑data training, PCSDiff cuts RMSE by 16.1\% and lifts ACC by 13.9\% relative to raw ECMWF forecasts at 3–10‑day lead times, and consistently outperforms mainstream deep‑learning baselines on both general and extreme‑precipitation metrics. Benefiting from a streaming inference pipeline, our method achieves low‑latency rolling forecasting for practical meteorological operations.
\end{abstract}

\begin{links}
    \link{Code}{https://github.com/zeaccepted/PCSDiff-Medium-term-precipitation-forecast}
\end{links}

\section{Introduction}
Weather forecasting has achieved great progress with conventional numerical models and modern AI methods, yet extreme precipitation prediction remains a challenging task \citep{bauer2015quiet}. Precipitation features strong nonlinear variations induced by complex terrain, convective microphysics and unstable monsoon circulations, leading to limited predictability especially for 10-day medium-range forecasts \citep{xiong2026csu}. Accurate medium-range precipitation correction is critical for national flood risk assessment, water resource management and drought-flood analysis over China \citep{liu2024multi}. Operational numerical forecasts, including GRAPES and ECMWF ensemble outputs, suffer from persistent systematic precipitation biases across China \citep{tan2025predictability, ye2014evaluation}. Such errors mainly stem from complex topographic gradients, East Asian summer monsoon variability, and insufficient model resolution for capturing migrating rainbands. Traditional statistical postprocessing methods, such as frequency matching and quantile mapping, only correct static biases and cannot model continuous spatiotemporal precipitation evolution. This causes cumulative forecast errors and degrades the reliability of operational medium-range precipitation forecasts \citep{ahmed2013statistical, maraun2016bias}.

In recent years, meteorological AI has developed diverse architectures for precipitation forecast correction. Static CNN-based methods such as U-Net \citep {ronneberger2015u} and U²-Net \citep {qin2020u2} refine single-frame precipitation spatial features. Spatiotemporal models including CNN-LSTM \citep {zhang2025deep}, SOM-CNN-LSTM \citep {zhang2023statistical}, and TransUNet \citep {ye2026dual} further capture temporal evolution of precipitation forecasts. Generative models have also advanced precipitation reconstruction tasks: GAN-based methods (Corrector-GAN \citep {price2022increasing}, SRGAN \citep {chen2022rainnet}) restore fine rainfall textures via adversarial optimization, while diffusion models (PrecipDiff \citep {dai2025precipdiff}, CorrDiff \citep {mardani2025residual}) recover subtle rainband details through iterative denoising. Additionally, emerging global meteorological foundation models, including FourCastNet \citep {pathak2022fourcastnet}, Pangu-Weather \citep {bi2023accurate}, GraphCast \citep {lam2023learning}, and AIFS \citep {lang2024aifs}, deliver powerful capabilities for medium-range atmospheric circulation prediction.

Despite the rapid progress of AI‑based precipitation correction methods, significant technical and practical limitations remain. Traditional calibration approaches cannot characterize daily rainband evolution, leading to error accumulation in medium‑range forecasting \citep{wu2026physical}. Most existing convolution, GAN, and vanilla diffusion models rely solely on general pixel‑level losses without meteorological constraints such as the Fraction Skill Score (FSS) \citep{mittermaier2010intercomparison}, frequently producing spatially unreasonable and over‑smoothed rainfall patterns. Meanwhile, general meteorological foundation models focus on full‑atmosphere autoregressive simulation and lack specialized designs for operational precipitation bias correction, struggling to preserve intense precipitation systems without dedicated multi‑day dynamic error modeling. Furthermore, current methods suffer from notable application drawbacks: most evaluations rely on discrete short‑term samples rather than continuous 10‑day medium‑range scenarios consistent with real operational services \citep{zhang2023statistical}, and few studies construct feasible deployment pipelines for migrating offline‑trained models to operational forecasting systems \citep{tang2023postrainbench}.

To address these issues, this work presents PCSDiff, a cascaded diffusion‑based two‑stage framework for 10‑day heavy precipitation bias correction and spatial downscaling over China. We pursue physical‑aware generalizable modeling by incorporating meteorological and terrain cues, alongside global‑dataset training and regional validation over China, to mitigate overfitting and stabilize performance for both conventional and extreme‑precipitation prediction. PCSDiff first adopts a Precipitation Intensity‑aware Multi‑branch Decoder (PIMD)‑based module for dynamic multi‑day precipitation bias correction under varied atmospheric and precipitation intensity conditions, and then employs a two‑phase super‑resolution module with conditional diffusion to reconstruct fine‑scale precipitation patterns. Trained on global meteorological data and validated on authoritative CMA observations, PCSDiff outperforms raw ECMWF forecasts and existing AI methods in numerical accuracy, temporal correlation, and spatial fidelity. Equipped with an optimized streaming inference pipeline, our model supports efficient and deployable forecasting for real operational scenarios. The main contributions are summarized as follows:
\begin{itemize}
\item We incorporate physical‑aware generalizable modeling by leveraging meteorological and terrain cues with global training and Chinese regional validation, achieving robust performance on conventional and extreme precipitation.
\item We propose a task‑decoupled cascaded framework with a PIMD‑based dynamic bias correction module and a conditional diffusion super‑resolution module, effectively mitigating multi‑day forecast biases and recovering fine‑scale precipitation details.
\item We implement a deployment‑efficient end‑to‑end streaming inference scheme. It enables low‑latency model deployment and supports practical operational precipitation forecasting workflows.
\end{itemize}

\section{Method: PCSDiff}
\begin{figure*}[t]
  \centering
  \includegraphics[width=0.95\textwidth]{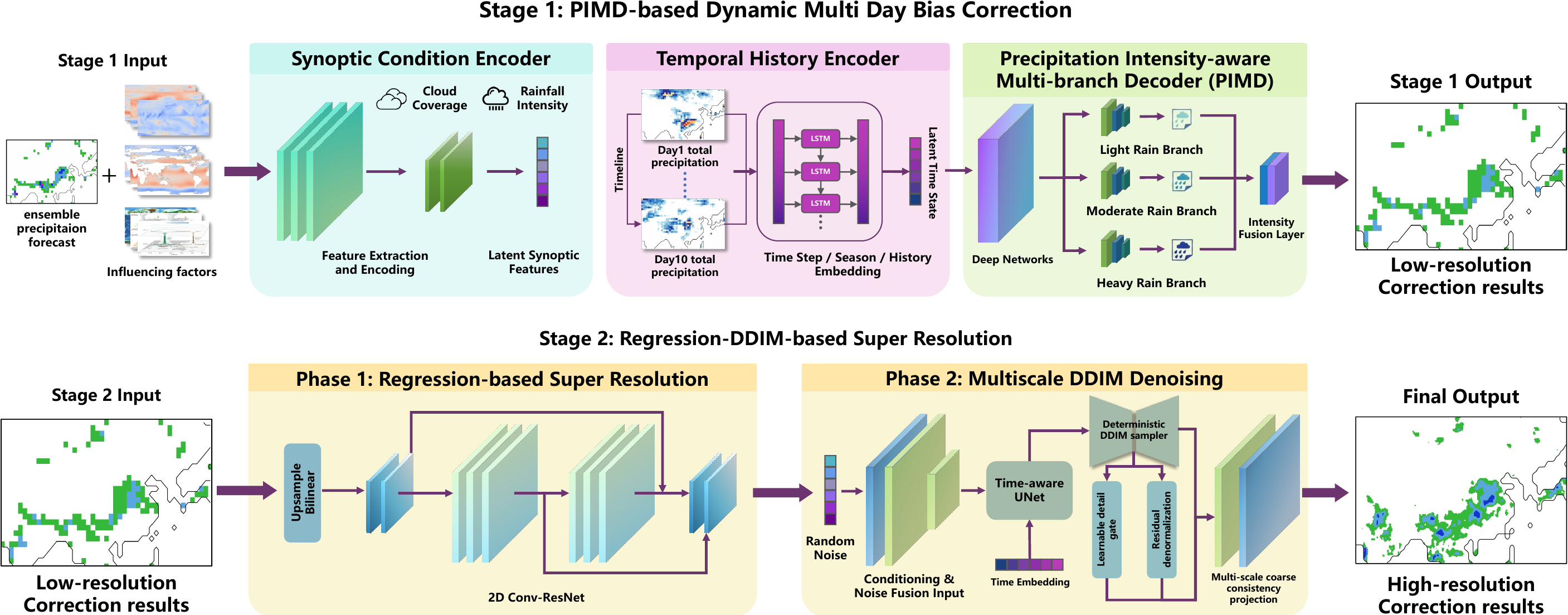}
  \caption{Overview of the proposed PCSDiff framework. The framework consists of two major stages. Stage 1 is the PIMD‑based dynamic multi‑day bias correction module, which takes 10‑day lead‑time ensemble precipitation forecasts as input, and employs the synoptic condition encoder, temporal history encoder, and precipitation intensity‑aware multi‑branch decoder (PIMD) to produce bias‑corrected low‑resolution precipitation. Stage 2 is the two‑phase super‑resolution module: Phase~1 applies a Conv‑ResNet for regression‑based preliminary upscaling, and Phase~2 further reconstructs fine‑scale precipitation details via a conditional denoising diffusion model with denoising U‑Net backbone, yielding the final high‑resolution corrected precipitation outputs.}
  \label{fig:fig1}
\end{figure*}

We target an end‑to‑end medium‑range precipitation bias‑correction and downscaling task: given coarse‑resolution 10‑day ensemble precipitation forecasts together with multi‑source physical covariates, our model outputs physically consistent high‑resolution precipitation fields.
We first define key notations:
\begin{itemize}
    \item $\mathcal{X}_{raw}\in\mathbb{R}^{T\times H\times W}$: raw coarse‑grid ensemble precipitation forecasts for $T$ forecast lead days, with spatial dimension $H\times W$;
    \item $\mathcal{F}_{syn}\in\mathbb{R}^{T\times H\times W\times C_s}$: physical‑aware synoptic and terrain auxiliary features, where $C_s$ denotes the number of meteorological and static terrain channels;
    \item $\mathcal{X}_{corr}\in\mathbb{R}^{T\times H\times W}$: low‑resolution precipitation after dynamic bias correction from Stage~1;
    \item $\hat{\mathcal{Y}}\in\mathbb{R}^{T\times H_h\times W_h}$: final high‑resolution corrected precipitation outputs, where $H_h,W_h$ denote high‑resolution spatial dimensions after downscaling ($H_h>H,\; W_h>W$).
\end{itemize}

As illustrated in Figure \ref{fig:fig1}, PCSDiff realizes this task via two sequential task‑decoupled stages, together with a deployment‑optimized inference pipeline. Guided by physical‑aware modeling principles, the whole framework ingests synoptic circulation and terrain priors to constrain meteorological plausibility for both bias correction and super‑resolution.
Stage~1 performs dynamic multi‑day bias correction:
\begin{equation}
\mathcal{X}_{corr} = \mathcal{M}_{\text{PIMD}}\big(\mathcal{X}_{raw},\mathcal{F}_{syn}\big)
\end{equation}
where $\mathcal{M}_{\text{PIMD}}$ denotes the PIMD‑based correction module. The subsequent two‑phase diffusion super‑resolution module maps corrected low‑resolution fields toward high‑resolution outputs:
\begin{equation}
\hat{\mathcal{Y}} = \mathcal{M}_{\text{SR}}\big(\mathcal{X}_{corr}\big)
\end{equation}
$\mathcal{M}_{\text{SR}}$ is composed of a regression‑based upscaling phase and a residual conditional DDIM diffusion phase, which preserves large‑scale meteorological patterns while reconstructing small‑scale convective details. Besides the two‑stage network forward pass, we further design a deployment‑oriented inference pipeline for practical operational weather prediction scenarios.

\subsubsection{Stage 1: PIMD-based Dynamic Multi-Day Bias Correction Module}
Given raw coarse‑resolution 10‑day ensemble precipitation $\mathcal{X}_{raw}$ and physical‑aware synoptic features $\mathcal{F}_{syn}$, this module $\mathcal{M}_{\text{PIMD}}$ aims to dynamically mitigate forecast biases evolving across lead times.
First, the \textit{Synoptic Condition Encoder} extracts latent synoptic representations from forecast covariates, encapsulating contextual information about cloud coverage and rainfall intensity. Second, the \textit{Temporal History Encoder} uses stacked LSTMs to model time‑series precipitation evolution over forecast lead days, producing time‑aware temporal embeddings that capture varying forecast states along the prediction timeline.
The synoptic latent features and temporal embeddings are fed into the \textit{Precipitation Intensity‑aware Multi‑branch Decoder (PIMD)}. PIMD adopts multi‑branch subnetworks specialized for light‑rain, moderate‑rain, and heavy‑rain regimes, to handle the heavily skewed statistical distribution of precipitation. An intensity fusion layer merges multi‑branch outputs and produces the bias‑corrected low‑resolution precipitation $\mathcal{X}_{corr}$ as the output of Stage~1.

\subsubsection{Stage 2: Regression‑DDIM‑based Super Resolution Module}
Directly generating full‑field high‑resolution precipitation with diffusion models risks corrupting large‑scale meteorological patterns embedded in $\mathcal{X}_{corr}$. To maintain physically reasonable large‑scale circulation features while recovering realistic small‑scale convective textures, we formulate $\mathcal{M}_{\text{SR}}$ as a two‑phase decomposition strategy. Phase~1 generates deterministic large‑scale precipitation priors, and Phase~2 conditional diffusion only learns residual fine‑scale details upon these priors, rather than generating complete high‑resolution fields from scratch.

In \textit{Phase 1: Regression‑based Super‑Resolution}, a 2‑D Conv‑ResNet implements deterministic upscaling regression:
\begin{equation}
\mathcal{Y}_{\text{prior}} = \mathcal{M}_{\text{ResNet}}(\mathcal{X}_{corr})
\end{equation}
$\mathcal{M}_{\text{ResNet}}$ learns the mapping from coarse inputs to intermediate high‑resolution representations $\mathcal{Y}_{\text{prior}}$, recovering large‑scale spatial structures and suppressing artifacts inherited from low‑resolution forecasts.

In \textit{Phase 2: Conditional Residual DDIM Diffusion}, we adopt a velocity‑prediction denoising U‑Net as the diffusion backbone. Taking concatenated outputs from Phase~1 as conditional input, the diffusion model learns only fine‑scale residual precipitation components $\Delta \mathcal{Y}$:
\begin{equation}
\Delta \mathcal{Y} = \mathcal{M}_{\text{DDIM}}\big(\mathcal{Y}_{\text{prior}}, z\big)
\end{equation}
where $z$ denotes random Gaussian noise. We adopt deterministic DDIM sampling with $\eta=0$ for fast inference to produce realistic small‑scale convective rainfall structures.
A composition fusion module with learnable detail gate and multi‑scale coarse‑grid consistency projection further protects large‑scale patterns from diffusion‑induced distortion. The final high‑resolution prediction is obtained by fusing prior fields and denoised residuals:
\begin{equation}
\hat{\mathcal{Y}} = \text{Fuse}\big(\mathcal{Y}_{\text{prior}},\Delta \mathcal{Y}\big)
\end{equation}
This fusion step enforces physical‑driven spatial consistency between coarse‑scale background circulation and reconstructed small‑scale rainfall details.

\subsubsection{Deployment-Oriented Inference Pipeline.}

Real‑world sub‑seasonal ensemble precipitation forecasting imposes strict constraints on inference efficiency, memory usage and operational compatibility, which are frequently neglected in research‑driven model designs. To bridge the gap between offline training and practical deployment, we build a unified inference pipeline for PCSDiff covering preprocessing, staged model execution, adaptive acceleration, and post‑processing. During inference, raw 10‑day lead‑time ensemble forecasts are fed as inputs. We adopt in‑memory tensor streaming: Stage 1 PIMD outputs are directly passed to the super‑resolution module without intermediate disk I/O, lowering storage overhead and end‑to‑end latency. For the conditional diffusion in Stage 2, sampling steps serve as a tunable trade‑off between generation quality and speed. Our pipeline supports dynamic sampling‑step configuration: full steps for high‑fidelity offline evaluation and accelerated sampling for time‑critical operational forecasting. We further implement lightweight pre‑/post‑processing tailored for ensemble precipitation inputs. Input normalization is computed on‑the‑fly without pre‑computed static datasets. After super‑resolution, physical‑constrained post‑processing clips non‑physical negative precipitation values while preserving ensemble statistical properties. The full pipeline can be packaged as a reusable module compatible with standard numerical‑weather‑prediction workflows and supports batch inference for multi‑member ensemble forecasts. Detailed information and performance are provided in the Appendix.

\section{Data and Experimental Setup}

\subsubsection{Data.}

This study employs long‑term operational forecasts and ground‑truth precipitation datasets spanning 2004‑2024, covering complete seasonal drought‑flood cycles over China to ensure sample diversity and model generalization. We use ECMWF global medium‑range forecasts (deterministic control and ensemble members) as model inputs. Besides total precipitation (tp), we incorporate physically relevant atmospheric variables and static terrain elevation to characterize moisture transport, ascending motion, large‑scale circulation and orographic precipitation effects. Full variable descriptions and detailed dataset specifications are provided in the Appendix. The raw ECMWF forecasts are at \(1.5^\circ\) horizontal resolution, while the reference ground truth is the 0.25° CMA‑CRA precipitation reanalysis dataset. The model is trained on global domains to learn universal precipitation spatiotemporal patterns and alleviate regional overfitting, and evaluations are confined to mainland China for 10‑day medium‑range heavy‑precipitation bias correction and downscaling.

\subsubsection{Training and Inference Configurations.}

We employ stage-specific training configurations for the different components of PCSDiff; full hyperparameters, training schedules, hardware specifications and implementation details are provided in the Appendix. To avoid temporal data leakage and mimic real‑world operational forecasting, the long‑term dataset is chronologically split into training and test sets, with the test set reserved solely for quantitative evaluation and ablation studies. PCSDiff is trained using a stage-wise optimization strategy. Stage 1 is optimized using a combination of MSE and TP losses, together with entropy regularization for the PIMD module. Stage 2 is trained in two phases: the regression-based super-resolution model is optimized using an intensity-weighted Smooth L1 loss with gradient regularization, while the conditional diffusion model is optimized using a composite diffusion objective. TP is our customized total‑precipitation score (see Appendix for definition), to jointly optimize precipitation bias correction and spatial super‑resolution downscaling. Detailed training configurations are provided in the Appendix.

\subsubsection{Evaluation Indicators.}

To comprehensively and quantitatively evaluate the model’s overall performance in terms of numerical precision, temporal correlation, spatial pattern rationality, and extreme precipitation prediction capability, we adopt standard meteorological verification systems and classify evaluation metrics into two major categories. Specifically, numerical error metrics including Root Mean Square Error (RMSE), Mean Absolute Error (MAE), Anomaly Correlation Coefficient (ACC) and TP are used to quantify the numerical accuracy and temporal correlation between model predictions and ground truth precipitation data. Meanwhile, extreme and spatial verification metrics covering Critical Success Index (CSI) and Fraction Skill Score (FSS) are employed to evaluate the model’s ability to identify extreme precipitation events and reconstruct realistic fine-scale spatial precipitation patterns. All mathematical definitions and detailed formulas of the above evaluation metrics are elaborated in the Appendix.

\section{Experiments and Results}
This section evaluates the proposed PCSDiff model for 10-day medium-range precipitation bias correction and downscaling over China, through quantitative comparison, qualitative event visualization, and module ablation analysis. We benchmark PCSDiff against raw ECMWF ensemble forecasts and mainstream data-driven methods. Core results including multi-metric quantitative evaluation, temporal performance variation, and typical heavy rainfall case verification are presented below, while additional sensitivity analyses and supplementary experiments are placed in the Appendix.

\subsubsection{Quantitative Result.}

\begin{figure*}[t]
  \centering
  \includegraphics[width=1\textwidth]{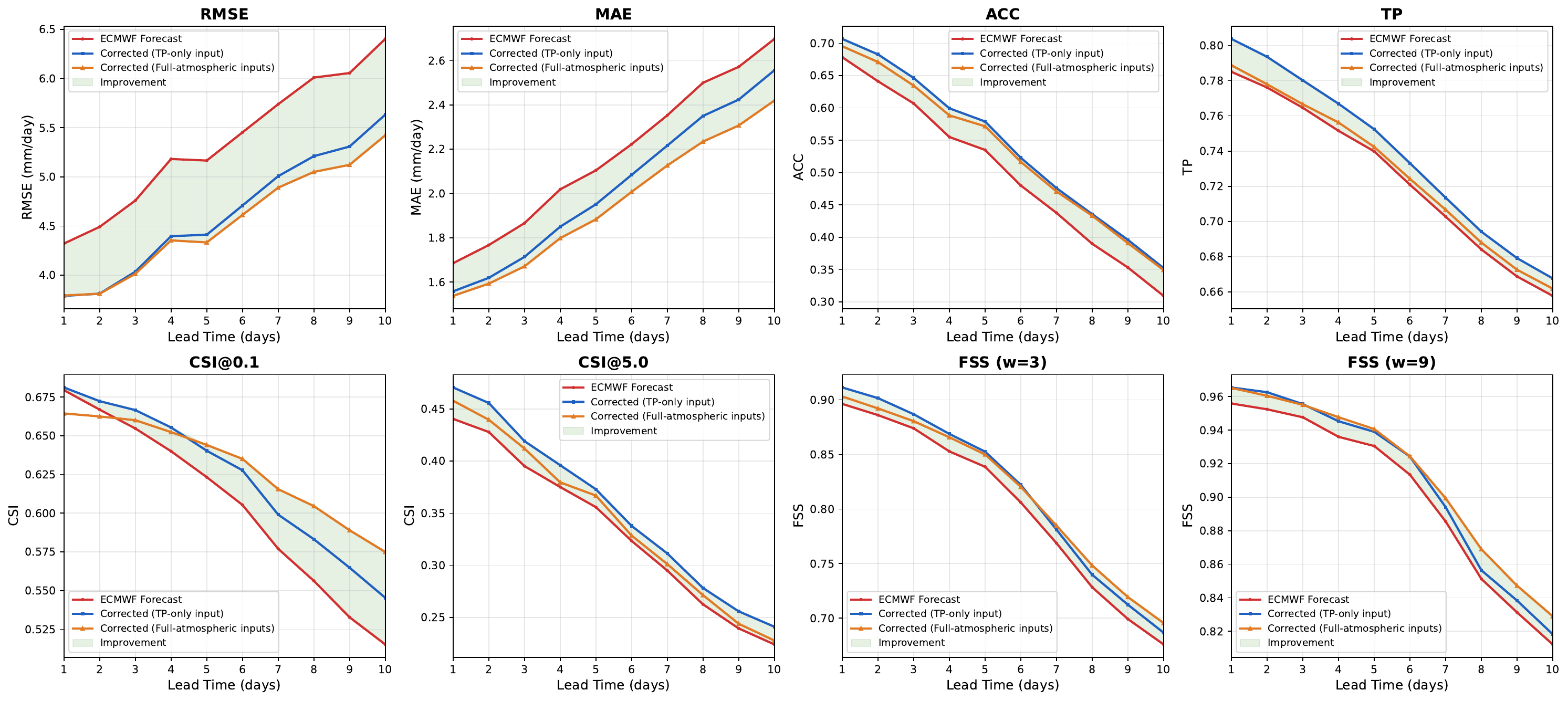}
  \caption{Performance metrics versus forecast lead time over China. Shown are raw ECMWF forecasts and two variants of our model using TP‑only and full‑atmospheric inputs. Subplots include continuous metrics (RMSE, MAE, ACC), event‑based TP and CSI scores and spatial FSS score. Light‑green shading denotes the performance gain of our model over raw ECMWF forecasts, evaluated on 1–10‑day forecasts over China.}
  \label{fig:metric_line}
\end{figure*}

\begin{table*}[t]
\centering
\caption{Performance comparison against baseline methods at 3‑day and 5‑day forecast lead times over China region}
\label{tab:main_result_35}
\begin{tabular}{l *{10}{c}}
\toprule
\multirow{2}{*}{Method}
& \multicolumn{5}{c}{Lead time = 3 day}
& \multicolumn{5}{c}{Lead time = 5 day} \\
\cmidrule(lr){2-6} \cmidrule(lr){7-11}
& RMSE$\downarrow$ & ACC$\uparrow$ & TP$\uparrow$ & CSI$\uparrow$ & FSS$\uparrow$
& RMSE$\downarrow$ & ACC$\uparrow$ & TP$\uparrow$ & CSI$\uparrow$ & FSS$\uparrow$ \\
\midrule
ECMWF
& 4.759 & 0.607 & 0.765 & 0.655 & 0.874
& 5.166 & 0.535 & 0.740 & 0.623 & 0.839 \\
ViT\citep {dosovitskiy2020image}
& 4.317 & 0.598 & 0.743 & 0.595 & 0.867
& 4.770 & 0.526 & 0.721 & 0.571 & 0.840 \\
ConvLSTM\cite{zhang2025deep}
& 4.405 & 0.583 & 0.729 & 0.587 & 0.851
& 4.807 & 0.519 & 0.707 & 0.565 & 0.824 \\
TransUNet\citep {ye2026dual}
& 4.344 & 0.593 & 0.731 & 0.579 & 0.851
& 4.782 & 0.524 & 0.711 & 0.558 & 0.828 \\
CorrDiff\citep {mardani2025residual}
& 4.404 & 0.589 & 0.737 & 0.602 & 0.855
& 4.830 & 0.522 & 0.715 & 0.572 & 0.829 \\
\midrule
\textbf{PCSDiff (Ours)}
& \textbf{4.014} & \textbf{0.646} & \textbf{0.780} & \textbf{0.667} & \textbf{0.887}
& \textbf{4.333} & \textbf{0.580} & \textbf{0.752} & \textbf{0.644} & \textbf{0.853} \\
\bottomrule
\end{tabular}
\end{table*}

We quantitatively assess PCSDiff’s numerical precision and temporal correlation at 3/5/7/10‑day lead times, with comprehensive performance reported in Table \ref{tab:main_result_35} and Table \ref{tab:main_result_710}. Though baselines (ViT, ConvLSTM, TransUNet, CorrDiff) lower continuous‑value RMSE, they suppress near‑threshold grids and over‑smooth rainfall structures, degrading threshold‑dependent metrics CSI, TP and FSS particularly for extreme precipitation. Compared with raw ECMWF outputs, PCSDiff reduces RMSE by 14.8\%–16.1\% and improves ACC by 6.4\%–13.9\%, yielding consistent gains across all lead times and mitigating medium‑range error accumulation. By introducing explicit constraints for precipitation event structures, PCSDiff boosts both continuous‑value and categorical metrics simultaneously, delivering superior CSI and FSS scores for heavy‑precipitation detection and reconstruction. Figure \ref{fig:metric_line} illustrates how PCSDiff’s performance evolves across different forecast lead times. PCSDiff retains stable, competitive predictive skill, demonstrating the efficacy of the PIMD‑driven dynamic bias correction and multi‑stage optimization.

\subsubsection{Qualitative Visualization of Typical Heavy Precipitation Events.}

\begin{figure*}[t]
  \centering
  \includegraphics[width=1\textwidth]{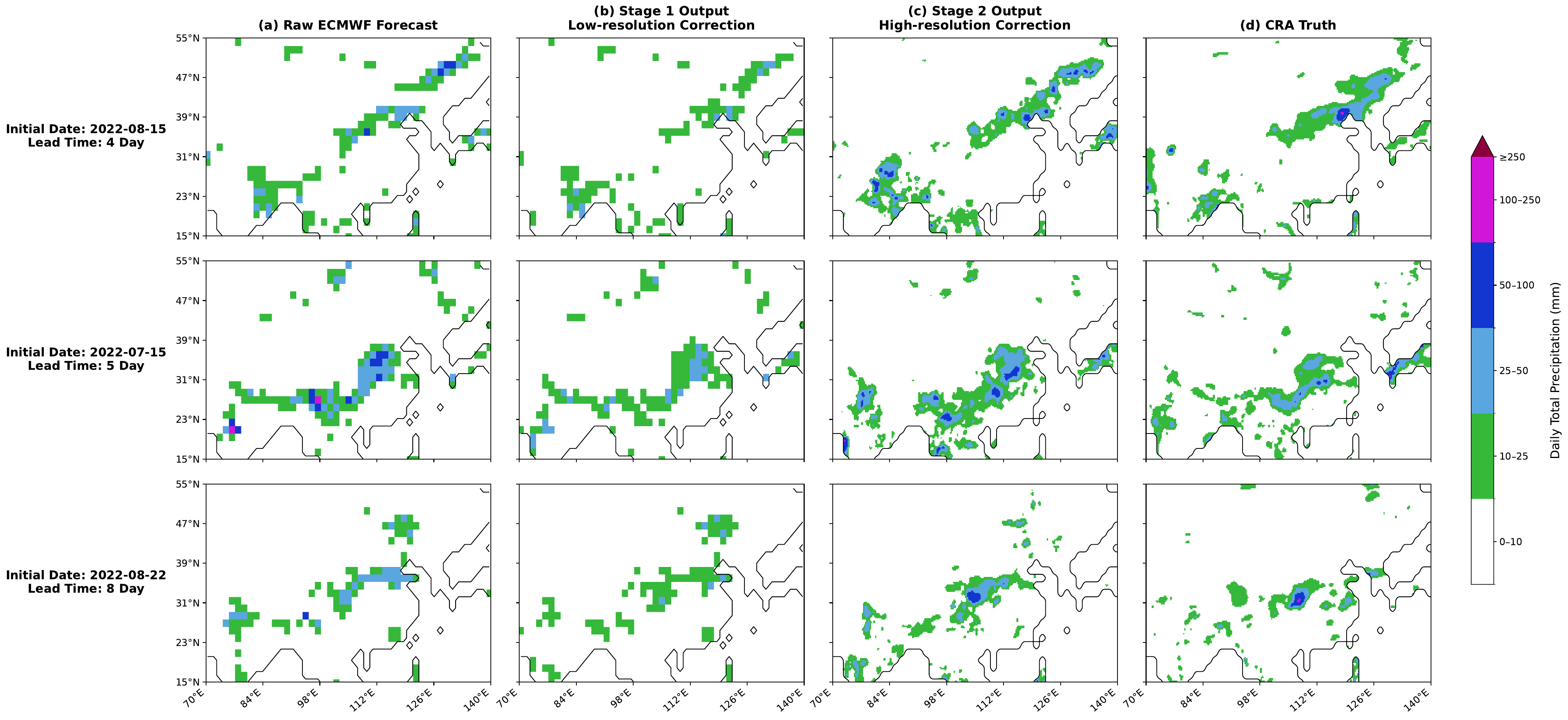}
  \caption{Spatial precipitation distributions for three representative heavy‑rainfall events. Each row corresponds to one case on different date and lead time. From left to right: raw ECMWF forecast, low‑resolution bias‑corrected output, high‑resolution super‑resolution result, and CMA‑CRA ground truth.}
  \label{fig:spatital_distribution}
\end{figure*}

To intuitively validate the spatial‑pattern reconstruction and fine‑detail restoration capability of PCSDiff, we visualize typical heavy‑rainfall events over China from 2022 under different forecast lead times. Figure \ref{fig:spatital_distribution} compares raw ECMWF forecasts, low‑resolution bias‑corrected outputs, PCSDiff high‑resolution super‑resolved predictions, and CMA‑CRA ground‑truth observations. Raw ECMWF coarse forecasts show blurred precipitation boundaries, missing small‑scale rain centers, and unrealistic spatial artifacts. The low‑resolution correction step mitigates large‑scale forecast biases, while the subsequent super‑resolution branch recovers fine‑scale orographic features and localized heavy‑rain structures. These case‑study examples across multiple lead‑time settings verify that our two‑phase pipeline yields physically plausible high‑resolution precipitation fields that closely match real observed rainfall patterns.

\subsubsection{Ablation Study.}

We conduct extensive ablation experiments to validate the effectiveness of key designs within PCSDiff. The ablated components include the PIMD module, loss‑function configurations, and combinations of input variables. Quantitative results for the different model variants are summarized in Table \ref{tab:ablation1} and \ref{tab:ablation2}. Experimental results demonstrate that discarding or modifying any of these critical designs yields clear performance drops in numerical accuracy, temporal correlation, and extreme‑precipitation detection capability. This confirms that the PIMD module, the adopted loss formulation, and the selected input variables are all essential, and their joint contribution enables the model to attain robust medium‑range precipitation forecasting. Further ablation studies on diffusion sampling steps, training strategies and hyper‑parameter settings are reported in the Appendix.

\section{Practical Deployment and Emerging Application Analysis}
Most existing meteorological‑AI precipitation correction models focus primarily on offline evaluation and case‑study validation, while largely overlooking real‑world engineering requirements for operational forecasting. Without standardized deployment pipelines and facing heavy computational overhead, such models can hardly sustain continuous rolling medium‑range forecasting, forming a notable gap between academic prototypes and meteorological operational services. Inspired by real‑world AI‑weather systems such as Google Weather AI, this section briefly analyzes PCSDiff’s deployability, practical strengths, and typical application scenarios. Comprehensive engineering implementation details and extended use‑case discussions are provided in the Appendix.

\subsubsection{Practical Deployment Barriers of Current Forecasting Models.}
Current AI precipitation forecasting methods face three major deployment bottlenecks: incompatibility with operational 10‑day rolling forecast workflows, low inference efficiency caused by non‑decoupled architectures, and the lack of complete workflows for porting offline‑trained models onto operational platforms.

\subsubsection{Deployment‑Oriented Pipeline of PCSDiff.}
Tailored for operational medium‑range precipitation forecasting, PCSDiff adopts a lightweight, portable cascaded architecture and supports the full engineering workflow from offline training to rolling batch inference. The in‑memory tensor‑streaming design removes intermediate disk I/O overhead. Benefiting from decoupled modules, PCSDiff requires one‑time pre‑training and supports low‑cost parallel inference for stable continuous precipitation product generation.

\subsubsection{Practical Application Value and Representative Use‑Cases.}
PCSDiff's high‑precision gridded precipitation products support these representative practical scenarios:
\begin{itemize}
    \item \textbf{Flood‑drought risk early warning}: Basin‑scale flood assessment, flash‑flood pre‑warning and seasonal drought monitoring for disaster‑management authorities.
    \item \textbf{Water resource management}: Reservoir operation scheduling and cross‑basin water‑resource allocation.
    \item \textbf{Public‑oriented weather services}: High‑resolution precipitation reference for end‑user‑facing weather applications.
\end{itemize}

\section{Conclusion and Future Work}

This paper presents PCSDiff, a task‑decoupled cascaded diffusion framework for 10‑day medium‑range precipitation bias correction and spatial downscaling over China. Equipped with a PIMD‑based dynamic multi‑day bias correction module and a two‑phase super‑resolution diffusion module, PCSDiff alleviates cumulative forecast errors and reconstructs physically consistent fine‑scale precipitation spatial patterns. Trained on global meteorological datasets and validated on authoritative CMA‑CRA observations, our model achieves competitive performance on general and extreme precipitation metrics against raw forecasts and mainstream AI baselines. Furthermore, the optimized streaming inference pipeline enables low‑latency migration to real‑world operational forecasting systems, supporting practical drought‑flood disaster prevention applications. 

In future, we will incorporate enhanced physical‑governed constraints within diffusion generation to mitigate physically implausible precipitation artifacts, extend the framework toward joint multi‑variable forecasting of temperature, humidity, and wind, and develop compressed lightweight model variants with accelerated sampling for resource‑constrained operational platforms. We also plan to explore domain‑adaptive fine‑tuning to boost performance for rare extreme weather events, and carry out real‑time offline trial runs in collaboration with meteorological agencies to advance the practical deployment of AI‑driven medium‑range forecasting systems.

\onecolumn
\newpage
\twocolumn
\section*{Acknowledgments}
This work was supported by the National Natural Science Foundation of China (Grant No. U2442206, 42125503, 42430602), Fundamental and Interdisciplinary Disciplines Breakthrough Plan of the Ministry of Education of China (Grant No. JYB2025XDXM801) and the China Meteorological Administration Youth Innovation Team (Grant No. CMA2024QN06).

\bigskip

\bibliography{aaai2027}

@article{bauer2015quiet,
  title={The quiet revolution of numerical weather prediction},
  author={Bauer, Peter and Thorpe, Alan and Brunet, Gilbert},
  journal={Nature},
  volume={525},
  number={7567},
  pages={47--55},
  year={2015},
  publisher={Nature Publishing Group UK London}
}

@article{xiong2026csu,
  title={CSU-PCAST: a dual-branch transformer framework for medium-range ensemble precipitation forecasting},
  author={Xiong, Tianyi and Chen, Haonan and Mahoney, Kelly and Tang, Jingyin and Smith, Tim and Bytheway, Janice},
  journal={npj Climate and Atmospheric Science},
  year={2026},
  publisher={Nature Publishing Group UK London}
}

@article{liu2024multi,
  title={Multi-model ensemble bias-corrected precipitation dataset and its application in identification of drought-flood abrupt alternation in China},
  author={Liu, Tingting and Zhu, Xiufang and Tang, Mingxiu and Guo, Chunhua and Lu, Dongyan},
  journal={Atmospheric Research},
  volume={307},
  pages={107481},
  year={2024},
  publisher={Elsevier}
}

@article{tan2025predictability,
  title={Predictability analysis based on ensemble forecasting of the “7{\textperiodcentered} 20” extreme rainstorm in Henan, China},
  author={Tan, Sai and Wang, Qiuping and Ma, Xulin and Sun, Lu and Zhang, Xin and Lv, Xinlu and Sun, Xin},
  journal={Frontiers of Earth Science},
  volume={19},
  number={3},
  pages={341--356},
  year={2025},
  publisher={Springer}
}

@article{ye2014evaluation,
  title={Evaluation of ECMWF medium-range ensemble forecasts of precipitation for river basins},
  author={Ye, Jinyin and He, Yi and Pappenberger, Florian and Cloke, Hannah L and Manful, Desmond Y and Li, Zhijia},
  journal={Quarterly Journal of the Royal Meteorological Society},
  volume={140},
  number={682},
  pages={1615--1628},
  year={2014},
  publisher={Wiley Online Library}
}

@article{ahmed2013statistical,
  title={Statistical downscaling and bias correction of climate model outputs for climate change impact assessment in the US northeast},
  author={Ahmed, Kazi Farzan and Wang, Guiling and Silander, John and Wilson, Adam M and Allen, Jenica M and Horton, Radley and Anyah, Richard},
  journal={Global and Planetary Change},
  volume={100},
  pages={320--332},
  year={2013},
  publisher={Elsevier}
}

@article{maraun2016bias,
  title={Bias correcting climate change simulations-a critical review},
  author={Maraun, Douglas},
  journal={Current Climate Change Reports},
  volume={2},
  number={4},
  pages={211--220},
  year={2016},
  publisher={Springer}
}

@inproceedings{ronneberger2015u,
  title={U-net: Convolutional networks for biomedical image segmentation},
  author={Ronneberger, Olaf and Fischer, Philipp and Brox, Thomas},
  booktitle={International Conference on Medical image computing and computer-assisted intervention},
  pages={234--241},
  year={2015},
  organization={Springer}
}

@article{qin2020u2,
  title={U2-Net: Going deeper with nested U-structure for salient object detection},
  author={Qin, Xuebin and Zhang, Zichen and Huang, Chenyang and Dehghan, Masood and Zaiane, Osmar R and Jagersand, Martin},
  journal={Pattern recognition},
  volume={106},
  pages={107404},
  year={2020},
  publisher={Elsevier}
}

@article{zhang2025deep,
  title={Deep learning for multi-source precipitation fusion on the Qinghai--Tibet Plateau},
  author={Zhang, Wenjuan and Di, Zhenhua},
  journal={International Journal of Digital Earth},
  volume={18},
  number={2},
  pages={2594247},
  year={2025},
  publisher={Taylor \& Francis}
}

@article{zhang2023statistical,
  title={Statistical post-processing of precipitation forecasts using circulation classifications and spatiotemporal deep neural networks},
  author={Zhang, Tuantuan and Liang, Zhongmin and Li, Wentao and Wang, Jun and Hu, Yiming and Li, Binquan},
  journal={Hydrology and Earth System Sciences},
  volume={27},
  number={10},
  pages={1945--1960},
  year={2023},
  publisher={Copernicus Publications G{\"o}ttingen, Germany}
}

@article{ye2026dual,
  title={A Dual-TransUNet Deep Learning Framework for Multi-Source Precipitation Merging and Improving Seasonal and Extreme Estimates},
  author={Ye, Yuchen and Qi, Zixuan and Li, Shixuan and Qi, Wei and Cai, Yanpeng and Yuan, Chaoxia},
  journal={arXiv preprint arXiv:2602.04757},
  year={2026}
}

@inproceedings{price2022increasing,
  title={Increasing the accuracy and resolution of precipitation forecasts using deep generative models},
  author={Price, Ilan and Rasp, Stephan},
  booktitle={International conference on artificial intelligence and statistics},
  pages={10555--10571},
  year={2022},
  organization={PMLR}
}

@article{chen2022rainnet,
  title={Rainnet: A large-scale imagery dataset and benchmark for spatial precipitation downscaling},
  author={Chen, Xuanhong and Feng, Kairui and Liu, Naiyuan and Ni, Bingbing and Lu, Yifan and Tong, Zhengyan and Liu, Ziang},
  journal={Advances in Neural Information Processing Systems},
  volume={35},
  pages={9797--9812},
  year={2022}
}

@inproceedings{dai2025precipdiff,
  title={PrecipDiff: Leveraging image diffusion models to enhance satellite-based precipitation observations},
  author={Dai, Ting-Yu and Ushijima-Mwesigwa, Hayato},
  booktitle={Proceedings of the AAAI Conference on Artificial Intelligence},
  volume={39},
  number={27},
  pages={27932--27939},
  year={2025}
}

@article{mardani2025residual,
  title={Residual corrective diffusion modeling for km-scale atmospheric downscaling},
  author={Mardani, Morteza and Brenowitz, Noah and Cohen, Yair and Pathak, Jaideep and Chen, Chieh-Yu and Liu, Cheng-Chin and Vahdat, Arash and Nabian, Mohammad Amin and Ge, Tao and Subramaniam, Akshay and others},
  journal={Communications Earth \& Environment},
  volume={6},
  number={1},
  pages={124},
  year={2025},
  publisher={Nature Publishing Group UK London}
}

@article{pathak2022fourcastnet,
  title={Fourcastnet: A global data-driven high-resolution weather model using adaptive fourier neural operators},
  author={Pathak, Jaideep and Subramanian, Shashank and Harrington, Peter and Raja, Sanjeev and Chattopadhyay, Ashesh and Mardani, Morteza and Kurth, Thorsten and Hall, David and Li, Zongyi and Azizzadenesheli, Kamyar and others},
  journal={arXiv preprint arXiv:2202.11214},
  year={2022}
}

@article{bi2023accurate,
  title={Accurate medium-range global weather forecasting with 3D neural networks},
  author={Bi, Kaifeng and Xie, Lingxi and Zhang, Hengheng and Chen, Xin and Gu, Xiaotao and Tian, Qi},
  journal={Nature},
  volume={619},
  number={7970},
  pages={533--538},
  year={2023},
  publisher={Nature Publishing Group UK London}
}

@article{lam2023learning,
  title={Learning skillful medium-range global weather forecasting},
  author={Lam, Remi and Sanchez-Gonzalez, Alvaro and Willson, Matthew and Wirnsberger, Peter and Fortunato, Meire and Alet, Ferran and Ravuri, Suman and Ewalds, Timo and Eaton-Rosen, Zach and Hu, Weihua and others},
  journal={Science},
  volume={382},
  number={6677},
  pages={1416--1421},
  year={2023},
  publisher={American Association for the Advancement of Science}
}

@article{lang2024aifs,
  title={AIFS--ECMWF's data-driven forecasting system},
  author={Lang, Simon and Alexe, Mihai and Chantry, Matthew and Dramsch, Jesper and Pinault, Florian and Raoult, Baudouin and Clare, Mariana CA and Lessig, Christian and Maier-Gerber, Michael and Magnusson, Linus and others},
  journal={arXiv preprint arXiv:2406.01465},
  year={2024}
}

@article{wu2026physical,
  title={Physical-Background-Constrained Bias Correction for Daily Precipitation Prediction Over China by a Subseasonal-to-Seasonal Model},
  author={Wu, Jie and Guo, Li and Jia, Xiaolong},
  journal={Atmospheric Science Letters},
  volume={27},
  number={3},
  pages={e70015},
  year={2026},
  publisher={Wiley Online Library}
}

@article{mittermaier2010intercomparison,
  title={Intercomparison of spatial forecast verification methods: Identifying skillful spatial scales using the fractions skill score},
  author={Mittermaier, Marion and Roberts, Nigel},
  journal={Weather and Forecasting},
  volume={25},
  number={1},
  pages={343--354},
  year={2010}
}

@article{tang2023postrainbench,
  title={Postrainbench: A comprehensive benchmark and a new model for precipitation forecasting},
  author={Tang, Yujin and Zhou, Jiaming and Pan, Xiang and Gong, Zeying and Liang, Junwei},
  journal={arXiv preprint arXiv:2310.02676},
  year={2023}
}

@article{dosovitskiy2020image,
  title={An image is worth 16x16 words: Transformers for image recognition at scale},
  author={Dosovitskiy, Alexey and Beyer, Lucas and Kolesnikov, Alexander and Weissenborn, Dirk and Zhai, Xiaohua and Unterthiner, Thomas and Dehghani, Mostafa and Minderer, Matthias and Heigold, Georg and Gelly, Sylvain and others},
  journal={arXiv preprint arXiv:2010.11929},
  year={2020}
}

\newpage

\onecolumn
\section{Appendix}

\renewcommand{\thefigure}{S\arabic{figure}} 
\renewcommand{\thetable}{S\arabic{table}} 
\setcounter{figure}{0} 
\setcounter{table}{0}  
\subsection{Table}

\begin{table}[h]
\centering
\caption{Performance comparison against baseline methods at 7‑day and 10‑day forecast lead times over China region}
\label{tab:main_result_710}
\begin{tabular}{l *{10}{c}}
\toprule
\multirow{2}{*}{Method}
& \multicolumn{5}{c}{Lead time = 7 day}
& \multicolumn{5}{c}{Lead time = 10 day} \\
\cmidrule(lr){2-6} \cmidrule(lr){7-11}
& RMSE$\downarrow$ & ACC$\uparrow$ & TP$\uparrow$ & CSI$\uparrow$ & FSS$\uparrow$
& RMSE$\downarrow$ & ACC$\uparrow$ & TP$\uparrow$ & CSI$\uparrow$ & FSS$\uparrow$ \\
\midrule
ECMWF
& 5.739 & 0.438 & 0.703 & 0.577 & 0.769
& 6.402 & 0.309 & 0.658 & 0.515 & 0.676 \\
ViT\citep {dosovitskiy2020image}
& 5.280 & 0.432 & 0.686 & 0.536 & 0.775
& 5.816 & 0.327 & 0.648 & 0.496 & \textbf{0.699} \\
ConvLSTM\cite{zhang2025deep}
& 5.309 & 0.422 & 0.677 & 0.533 & 0.764
& 5.848 & 0.320 & 0.643 & 0.503 & 0.693 \\
TransUnet\citep {ye2026dual}
& 5.292 & 0.427 & 0.679 & 0.526 & 0.765
& 5.828 & 0.323 & 0.643 & 0.488 & 0.694 \\
CorrDiff\citep {mardani2025residual}
& 5.325 & 0.422 & 0.681 & 0.527 & 0.761
& 5.821 & 0.317 & 0.644 & 0.478 & 0.686 \\
\midrule
\textbf{PCSDiff (Ours)}
& \textbf{4.890} & \textbf{0.476} & \textbf{0.714} & \textbf{0.616} & \textbf{0.785}
& \textbf{5.423} & \textbf{0.352} & \textbf{0.668} & \textbf{0.575} & 0.696 \\
\bottomrule
\end{tabular}
\end{table}

\begin{table}[!htbp]
\centering
\caption{Input meteorological variables from ECMWF ensemble forecast.}
\label{tab:input_vars}
\begin{tabular}{l l l p{5.5cm}}
\toprule
Variable & Name & Pressure Levels & Role for total precipitation (tp) \\
\midrule
2t & 2m temperature & Single surface level & Determines precipitation phase \\
gh & geopotential height & 850hPa, 700hPa, 500hPa & Characterizes circulation patterns \\
msl & mean sea‑level pressure & Single surface level & Important for pressure systems and fronts \\
q & specific humidity & 1000hPa, 925hPa, 850hPa, 700hPa, 500hPa & Represents atmospheric moisture content \\
sp & surface pressure & Single surface level & Important surface boundary condition \\
sst & sea surface temperature & Single surface level & Significant over ocean regions \\
tp & total precipitation & Single surface level & Target forecast variable \\
u & zonal wind & 1000hPa, 925hPa, 850hPa, 700hPa, 500hPa, 300hPa & Moisture transport and convergence \\
v & meridional wind & 1000hPa, 925hPa, 850hPa, 700hPa, 500hPa, 300hPa & Cooperates with u for moisture advection \\
w & vertical velocity & 1000hPa, 925hPa, 850hPa, 700hPa, 500hPa, 300hPa & Ascending motion drives precipitation \\
\bottomrule
\end{tabular}
\end{table}

\begin{table}[h]
\centering
\caption{Ablation study of key components in PCSDiff over China region (Lead time = 3/5 day)}
\label{tab:ablation1}
\begin{tabular}{l *{10}{c}}
\toprule
\multirow{2}{*}{Method}
& \multicolumn{5}{c}{Lead time = 3 day}
& \multicolumn{5}{c}{Lead time = 5 day} \\
\cmidrule(lr){2-6} \cmidrule(lr){7-11}
& RMSE$\downarrow$ & ACC$\uparrow$ & TP$\uparrow$ & CSI$\uparrow$ & FSS$\uparrow$
& RMSE$\downarrow$ & ACC$\uparrow$ & TP$\uparrow$ & CSI$\uparrow$ & FSS$\uparrow$ \\
\midrule
ECMWF
& 4.759 & 0.607 & 0.765 & 0.655 & 0.874
& 5.166 & 0.535 & 0.740 & 0.623 & 0.839 \\
w/o other variables (only precipitation)
& 4.036 & 0.646 & 0.780 & 0.667 & 0.887
& 4.412 & 0.579 & 0.752 & 0.640 & 0.853 \\
w/o TP loss (MSE Loss)
& 4.220 & 0.627 & 0.758 & 0.644 & 0.871
& 4.531 & 0.562 & 0.734 & 0.628 & 0.838 \\
w/o TP loss (MSE + FSS Loss)
& 4.228 & 0.631 & 0.767 & 0.651 & 0.868
& 4.547 & 0.568 & 0.741 & 0.633 & 0.841 \\
w/o PIMD module
& 4.190 & 0.638 & 0.769 & 0.653 & 0.872
& 4.516 & 0.574 & 0.743 & 0.636 & 0.840 \\
\midrule
\textbf{PCSDiff (Ours, full setting)}
& \textbf{4.014} & \textbf{0.646} & \textbf{0.780} & \textbf{0.667} & \textbf{0.887}
& \textbf{4.333} & \textbf{0.580} & \textbf{0.752} & \textbf{0.644} & \textbf{0.853} \\
\bottomrule
\end{tabular}
\end{table}

\begin{table}[h]
\centering
\caption{Ablation study of key components in PCSDiff over China region (Lead time = 7/10 day)}
\label{tab:ablation2}
\begin{tabular}{l *{10}{c}}
\toprule
\multirow{2}{*}{Method}
& \multicolumn{5}{c}{Lead time = 7 day}
& \multicolumn{5}{c}{Lead time = 10 day} \\
\cmidrule(lr){2-6} \cmidrule(lr){7-11}
& RMSE$\downarrow$ & ACC$\uparrow$ & TP$\uparrow$ & CSI$\uparrow$ & FSS$\uparrow$
& RMSE$\downarrow$ & ACC$\uparrow$ & TP$\uparrow$ & CSI$\uparrow$ & FSS$\uparrow$ \\
\midrule
ECMWF
& 5.739 & 0.438 & 0.703 & 0.577 & 0.769
& 6.402 & 0.309 & 0.658 & 0.515 & 0.676 \\
w/o other variables (only precipitation)
& 5.007 & 0.476 & 0.714 & 0.599 & 0.781
& 5.632 & 0.352 & 0.668 & 0.545 & 0.687 \\
w/o TP loss (MSE Loss)
& 5.071 & 0.462 & 0.699 & 0.599 & 0.774
& 5.600 & 0.339 & 0.656 & 0.562 & 0.685 \\
w/o TP loss (MSE + FSS Loss)
& 5.099 & 0.468 & 0.705 & 0.601 & 0.778
& 5.642 & 0.345 & 0.660 & 0.562 & 0.689 \\
w/o PIMD module
& 5.074 & 0.472 & 0.707 & 0.596 & 0.775
& 5.613 & 0.348 & 0.662 & 0.559 & 0.683 \\
\midrule
\textbf{PCSDiff (Ours, full setting)}
& \textbf{4.890} & \textbf{0.476} & \textbf{0.714} & \textbf{0.616} & \textbf{0.785}
& \textbf{5.423} & \textbf{0.352} & \textbf{0.668} & \textbf{0.575} & \textbf{0.696} \\
\bottomrule
\end{tabular}
\end{table}

\begin{table}[!htbp]
\suppressfloats[t]
\centering
\caption{Training -- Stage~1 Hyper‑parameters.}
\label{tab:training_config1}
\begin{tabular}{l l l p{7cm}}
\toprule
Category & Parameter & Value & Description \\
\midrule
\multirow{26}{2cm}{\textbf{Training -- Stage~1}} 
& Input channels & 39 & 10 ensemble precipitation channels and 29 auxiliary atmospheric-variable channels \\
& Output channels & 1 & \\
& PIMD branches & 3 & Number of parallel PIMD branches \\
& Base feature dimension & 64 & Initial feature dimension of the U-Net \\
& Dimension multipliers & (1, 2, 4, 8) & Feature multipliers at different network levels \\
& Batch size per GPU & 1 & \\
& Number of GPUs & 2 & Distributed training using two GPUs \\
& Gradient accumulation & 4 & Number of gradient-accumulation steps\\
& Effective batch size & 8 & \(1\times2\times4=8\) \\
& Training iterations & 5000 & Total number of training iterations \\
& Optimizer updates & 1250 & One update per four accumulated iterations \\
& Learning rate & $1\times10^{-4}$ & \\
& Optimizer & AdamW & Optimizer with decoupled weight decay \\
& Weight decay & $1\times10^{-2}$ & \\
& Learning rate scheduler & Delayed cosine decay & Cosine decay activated after iteration 3000 \\
& Gradient clipping & 0.5 & Threshold for gradient norm clipping \\
& Early stopping & Not used & Convergence controlled by fixed update schedule \\
& Random seed & 42 & \\
& Loss function & MSE + TP Loss & Loss for deterministic precipitation correction \\
& TP-loss weight & 0.01 & \\
& TP thresholds & 0.1, 1.0, and 5.0 mm & Precipitation thresholds used in TP Loss \\
& PIMD regularization & Entropy regularization & Auxiliary regularization of PIMD branch weights \\
& PIMD regularization weight & 0.01 & Weight of the entropy-regularization term \\
& EMA decay & 0.995 & \\
& EMA start & Iteration 500 & \\
& EMA update interval & Every 10 iterations & \\
\bottomrule
\end{tabular}
\end{table}

\begin{table}[!htbp]
\suppressfloats[t]
\centering
\caption{Training -- Stage~2 Hyper‑parameters.}
\label{tab:training_config2}
\begin{tabular}{p{2cm} l p{4cm} p{6cm}}
\toprule
Category & Parameter & Value & Description \\
\midrule
\multirow{14}{2cm}{\textbf{Training -- Stage~2 Phase~1}} 
& Input channels & 1 &  \\
& Output channels & 1 & \\
& Base feature channels & 64 & Base feature width of the regression network \\
& Residual blocks & 8 & \\
& Residual scaling & 0.5 & Residual-feature scaling factor \\
& Batch size & 4 & \\
& Training epochs & 40 & \\
& Learning rate & \(2\times10^{-4}\) & \\
& Optimizer & AdamW & \\
& Weight decay & \(1\times10^{-4}\) & \\
& Gradient clipping & 1.0 & \\
& Early stopping & Not used & \\
& Loss function & Weighted Smooth L1 + gradient loss & Intensity-aware regression objective \\
& Gradient-loss weight & 0.05 & \\

\midrule
\multirow{19}{2cm}{\textbf{Training -- Stage~2 Phase~2}} 
& Denoiser input channels & 3 & One noisy residual and two conditioning channels \\
& Output channels & 1 & \\
& Base feature channels & 48 & Base feature width of the diffusion model \\
& Diffusion timesteps & 100 & \\
& DDIM sampling steps & 25 & \\
& Batch size & 4 & \\
& Maximum training epochs & 40 & \\
& Learning rate & \(1\times10^{-4}\) & \\
& Optimizer & AdamW & \\
& Weight decay & \(1\times10^{-4}\) & \\
& Gradient clipping & Not used & \\
& Early stopping & Patience = 8 & Based on end-to-end validation score \\
& Loss function & Composite diffusion loss & Velocity, reconstruction, consistency, gradient, structural, FSS, and gate losses \\
& Loss-component weights & \(1.0,\ 4.0,\ 0.1,\ 0.15,\ 0.8, \allowbreak\ 1.0,\ 0.5\) & Weights for the \(x_0\), reconstruction, consistency, gradient, structural, FSS, and gate losses, respectively \\
& Heavy-rain weight & 6.0 & Additional weight for heavy-precipitation regions \\
\bottomrule
\end{tabular}
\end{table}

\begin{table}[!htbp]
\suppressfloats[t]
\centering
\caption{Dataset and Hardware Configuration.}
\label{tab:training_config3}
\begin{tabular}{l l l p{7cm}}
\toprule
Category & Parameter & Value & Description \\
\midrule
\multirow{11}{*}{\textbf{Dataset}} 
& Train set & 2004–2021 & \\
& Test set & 2022–2024 & \\
& Forecast data & ECMWF S2S & Ensemble forecasts and atmospheric variables \\
& Reference precipitation & CMA-CRA & High-resolution precipitation observations \\
& Stage-1 resolution & \(1.5^\circ\rightarrow1.5^\circ\) & Forecast correction at the original resolution \\
& Stage-2 resolution & \(1.5^\circ\rightarrow0.25^\circ\) & Downscaling from ECMWF forecasts to the CMA-CRA resolution\\
& Input lead time & 1–10 days & \\
& Spatial domain & 15–55°N, 70–140°E & China and surrounding region \\
& Stage-1 normalization & Min‑max & Normalization based on training-set statistics \\
& Stage-2 normalization & Logarithmic scaling & Precipitation normalization using a 250-mm reference maximum \\
& Anomaly processing & Pentad climatology & Anomaly computation based on pentad‑scale climatology \\
& Land mask & Applied & Land-region masking where applicable \\

\midrule
\multirow{4}{*}{\textbf{Hardware}} 
& GPU & NVIDIA A100 PCIe 40 GB& \\
& CPU & Kunpeng 920& \\
& Stage-1 precision & Mixed precision & PyTorch autocast enabled without gradient scaling \\
& Stage-2 precision & FP32 & Mixed-precision training not used \\
\bottomrule
\end{tabular}
\end{table}

\newpage
\onecolumn

\subsection{Evaluation Metrics}
This appendix provides mathematical definitions for all evaluation metrics used in this work.
Let $f_{i,j}$ denote forecast precipitation at grid $(i,j)$, $x_{i,j}$ denote the corresponding ground‑truth precipitation, $N$ be the total number of valid grid points, and $j$ index forecast lead days ranging from $1$ to $10$.

\paragraph{RMSE (Root Mean Square Error)}
\[
\text{RMSE}=\sqrt{\frac{1}{N}\sum_{i,j}\big(f_{i,j}-x_{i,j}\big)^2}
\]
RMSE quantifies the overall magnitude of forecast error. Lower RMSE indicates better numerical accuracy.

\paragraph{MAE (Mean Absolute Error)}
\[
\text{MAE}=\frac{1}{N}\sum_{i,j}\big|f_{i,j}-x_{i,j}\big|
\]
MAE measures the mean absolute deviation between forecasts and observations. Smaller MAE corresponds to more accurate precipitation magnitude.

\paragraph{ACC (Anomaly Correlation Coefficient)}
\[
\text{ACC}=
\frac{\sum_{i,j}(f_{i,j}-\bar f)(x_{i,j}-\bar x)}
{\sqrt{\sum_{i,j}(f_{i,j}-\bar f)^2}\sqrt{\sum_{i,j}(x_{i,j}-\bar x)^2}}
\]
where $\bar f$ and $\bar x$ represent climatological mean of forecast and reference precipitation, respectively. ACC reflects the temporal correlation of precipitation anomalies; higher ACC implies better consistency of precipitation variation.

\paragraph{TP (Comprehensive total‑precipitation score, custom metric)}
The TP metric relies on intermediate categorical verification quantities defined below.
The contingency table is constructed by thresholding observed precipitation $x_{i,j}$ and forecast precipitation $f_{i,j}$ at $1\,\text{mm/day}$.
\begin{table}[h]
\centering
\caption{Contingency table for categorical precipitation verification (threshold: $1\,\text{mm/day}$)}
\begin{tabular}{lcc}
\toprule
Observation ($x_{i,j}$) & \multicolumn{2}{c}{Forecast ($f_{i,j}$)} \\
\cmidrule(lr){2-3}
 & $\ge 1\,\text{mm/day}$ & $< 1\,\text{mm/day}$ \\
\midrule
$\ge 1\,\text{mm/day}$ & NA (Hit) & NC (Miss) \\
$< 1\,\text{mm/day}$ & NB (False alarm) & ND (Correct negative) \\
\bottomrule
\end{tabular}
\end{table}
Definitions:
\begin{itemize}
    \item NA: Hit, both observation and forecast exceed the threshold.
    \item NC: Miss, observation exceeds threshold while forecast does not.
    \item NB: False alarm, forecast exceeds threshold while observation does not.
    \item ND: Correct negative, both observation and forecast are below the threshold.
\end{itemize}

For each forecast lead day $j$:
\[
TS_{1,j}=\frac{NA}{NA+NB+NC}
\]
$TS_{1,j}$ is the threat score evaluated at the $1\,\text{mm/day}$ threshold for general rainfall events.

The heavy‑precipitation threshold for each grid is defined as one standard deviation of local precipitation climatology. If the computed threshold is less than $3\,\text{mm/day}$, it is clamped to $3\,\text{mm/day}$. Under this spatially‑varying heavy‑rain threshold, the heavy‑rain threat score is:
\[
TS_{S,j}=\frac{NA_S}{NA_S+NB_S+NC_S}
\]
where $NA_S$, $NB_S$, $NC_S$ are hit, false‑alarm and miss counts evaluated with the heavy‑precipitation threshold. $TS_{S,j}$ assigns extra reward for well‑predicted heavy precipitation regions.

Synthetic threat score combines general‑rain and heavy‑rain performance:
\[
TS_{syn,j}=\frac{TS_{1,j}+TS_{S,j}}{1+TS_{S,j}}
\]

Proportion Correct:
\[
PC_j=\frac{NA+ND}{NA+NB+NC+ND}
\]
$PC_j$ denotes the overall categorical correct rate for lead day $j$.

Let $w_j$ be the weight of lead day $j$. The final comprehensive TP score aggregates results over lead days $j=1$ to $j=10$:
\[
TP=
\frac{\sum_{j=1}^{10} w_j \cdot \big(TS_{syn,j}+PC_j\big)/2}{\sum_{j=1}^{10} w_j}
\]
TP jointly considers precipitation hit skill and overall categorical accuracy. Larger TP indicates better forecast performance.

\paragraph{CSI (Critical Success Index)}
\[
\text{CSI} = \frac{NA}{NA+NB+NC}
\]
CSI (also known as Threat Score) measures categorical forecast skill for precipitation events. Higher CSI represents better categorical prediction skill.

\paragraph{FSS (Fraction Skill Score)}
\[
\text{FSS}=
1-\frac{\sum_{i,j}\big(O_{i,j}(n)-M_{i,j}(n)\big)^2}{\sum_{i,j} O_{i,j}(n)^2+\sum_{i,j} M_{i,j}(n)^2}
\]
where $O_{i,j}(n)$ and $M_{i,j}(n)$ are fractional coverage of precipitation exceeding a given threshold within an $n\times n$ local neighbourhood window for observation and forecast, respectively. FSS assesses spatial pattern similarity of precipitation fields and penalizes spatial displacement of rainbands. Its value ranges from $0$ to $1$, and higher FSS represents more realistic spatial precipitation structures.

All metrics are computed on individual grid points and then aggregated over the Chinese mainland domain.

\newpage
\onecolumn
\subsection{Implementation Details of the Deployment‑Oriented Inference Pipeline}
This appendix supplements concrete implementation configurations, hardware environment, and performance test details for the PCSDiff streaming inference pipeline, which is briefly described in the main text.

\subsubsection{Hardware and Software Environment}
All inference performance tests are conducted on a single consumer‑grade GPU. The test platform adopts NVIDIA RTX 4090 / A100 GPU (40 GB video memory), with PyTorch 2.1 as the deep‑learning framework. CUDA 12.1 is used for GPU acceleration. The model inference adopts FP32 precision by default; FP16 mixed‑precision can be optionally enabled for further memory reduction.

\subsubsection{Inference Pipeline Configuration}
The unified inference pipeline consists of four components: data preprocessing, Stage‑1 PIMD bias correction, Stage‑2 two‑phase diffusion super‑resolution, and physical‑constrained post‑processing.
\begin{enumerate}
    \item \textbf{Preprocessing}: Input normalization is computed on‑the‑fly according to forecast metadata. No offline pre‑computed normalization files are stored on disk. Input variables include ECMWF precipitation and atmospheric synoptic features together with static terrain elevation, consistent with the training phase.
    \item \textbf{In‑memory tensor streaming}: Output tensors from the PIMD module are directly passed to the subsequent super‑resolution module within GPU memory. Intermediate results are not serialized or written to disk, eliminating disk read‑write overhead.
    \item \textbf{Dynamic DDIM sampling steps}: Two sampling modes are supported. For offline high‑fidelity evaluation, we adopt full sampling steps ($N_{\text{step}}=50$). For operational accelerated inference, we reduce sampling steps ($N_{\text{step}}=10$) to trade minor spatial detail degradation for lower latency. The DDIM sampler is configured with $\eta=0$ for deterministic generation.
    \item \textbf{Post‑processing}: Non‑physical negative precipitation values are clipped to zero. A statistical rescaling operation is applied to preserve the mean and variance statistics of input ensemble precipitation forecasts.
\end{enumerate}

\subsubsection{Performance Test Settings}
The test case follows real operational input format: input corresponds to ECMWF 10‑day lead‑time ensemble forecast, with the input spatial resolution of $1.5^\circ$. The output precipitation field is downscaled to $0.25^\circ$ resolution. We test single‑member and multi‑member ensemble batch inference scenarios. Test samples are selected from test set (years 2022–2024), which is completely separated from the training dataset to avoid data leakage.

\subsubsection{Inference Latency and Memory Footprint}
Table \ref{tab:inference_performance} summarizes measured end‑to‑end inference wall‑clock time and GPU memory consumption under different sampling‑step configurations and different ensemble member numbers. Latency counts the full procedure from raw forecast input to final post‑processed high‑resolution precipitation output.

\begin{table}[h!]
\centering
\caption{Inference performance of PCSDiff deployment pipeline under different settings. Latency denotes wall‑clock time for complete 10‑day forecast inference.}
\label{tab:inference_performance}
\begin{tabular}{lccc}
\hline
Configuration & Ensemble Members & GPU Memory (GB) & End‑to‑end Latency (s) \\
\hline
Full sampling ($N_\text{step}=50$) & 5 & 12.4 & 14.7 \\
Full sampling ($N_\text{step}=50$) & 10 & 18.7 & 38.2 \\
Accelerated sampling ($N_\text{step}=10$) & 5 & 11.8 & 3.6 \\
Accelerated sampling ($N_\text{step}=10$) & 10 & 17.2 & 13.1 \\
\hline
\end{tabular}
\end{table}

From Table~\ref{tab:inference_performance}, in‑memory tensor streaming avoids frequent disk I/O operations and significantly cuts storage overhead. Accelerated sampling greatly reduces inference time with only limited degradation of spatial forecast skill. This characteristic enables PCSDiff to adapt to two usage modes: high‑quality offline research evaluation and low‑latency real‑time operational forecasting.

\subsubsection{Compatibility with NWP operational workflows}
The whole inference module is encapsulated as a Python callable class, decoupled from training logic. Input and output adopt standard geospatial array formats compatible with mainstream numerical weather prediction workflow tools. It supports continuous rolling inference for successive forecast initialization times.

\newpage
\onecolumn

\subsection{Additional Details on Model Deployment and Application Analysis}
This section supplements comprehensive engineering details of the PCSDiff deployment pipeline, quantitative inference performance analysis, extended application scenarios, and comparison against real‑world AI‑driven weather systems, which are briefly summarized in the main text.

\subsubsection{Detailed Analysis of Deployment Bottlenecks}
Three bottlenecks restricting real‑world deployment of existing AI precipitation correction models are elaborated as follows.
First, most existing models are validated on discrete short‑term samples rather than continuous 10‑day rolling forecast sequences. When migrated to operational workflows, they cannot continuously suppress lead‑time‑dependent error accumulation along multi‑day forecast horizons.
Second, many complex meteorological AI models adopt monolithic architectures without task‑decoupled design. Bias correction and super‑resolution computations are tightly coupled, bringing redundant computation and limiting parallel acceleration, which cannot satisfy the low‑latency requirements of real‑time operational forecasting.
Third, most research works only release model weights and training codes, lacking standardized pre‑processing, post‑processing, and model‑migration toolchains. Extra heavy engineering adaptation is required to port offline‑trained models onto operational meteorological platforms, which greatly hinders large‑scale practical promotion.

\subsubsection{Full Workflow of the PCSDiff Deployment‑Oriented Pipeline}
The complete inference pipeline consists of four sequential components: data preprocessing, stage‑1 bias correction, stage‑2 diffusion super‑resolution, and physical‑constrained post‑processing.
Instead of saving intermediate results to disk, the in‑memory tensor‑streaming mechanism directly transfers output tensors from the PIMD bias‑correction module to the subsequent super‑resolution module within GPU memory, eliminating disk read‑write latency.
For the diffusion super‑resolution branch, configurable DDIM sampling steps provide a flexibility trade‑off: high‑sampling‑step mode is used for offline high‑fidelity evaluation, while reduced‑step accelerated sampling is activated for time‑critical operational forecasting. Physical‑constrained post‑processing clamps non‑physical negative precipitation values and preserves the statistical characteristics of input ensemble forecasts.
Detailed measured inference latency under different GPU hardware and different sampling‑step configurations are listed in Table S9. PCSDiff supports batch processing for multi‑member ensemble forecasts and can be encapsulated as reusable components compatible with mainstream numerical weather prediction workflow.

\subsubsection{Extended Practical and Emerging Application Scenarios}
Beyond the three main use‑cases described in the main text, PCSDiff‑generated high‑resolution precipitation products can be further extended to additional application domains:
\begin{itemize}
    \item \textbf{Hydrological model driving data}: Serve as forcing inputs for distributed hydrological models to improve streamflow simulation accuracy for small‑and‑medium‑size catchments.
    \item \textbf{Agricultural meteorological services}: Support crop‑growth‑related meteorological assessment and agricultural drought monitoring at fine spatial scales.
    \item \textbf{Climate impact research}: Provide bias‑corrected high‑resolution medium‑range precipitation datasets for regional climate‑impact and risk‑related research.
\end{itemize}

\subsubsection{Comparison with Real‑World AI‑Enhanced Weather Systems}
Commercial AI‑weather services such as Google Weather AI produce high‑resolution gridded weather products for public end‑users. Such systems focus on delivering user‑oriented weather information, while their internal bias‑correction and downscaling implementation details are not publicly disclosed.
Different from closed commercial systems, PCSDiff is an open‑research‑oriented framework targeting medium‑range ensemble precipitation post‑processing. It provides transparent network design, complete training and inference pipeline, and supports secondary development for domain‑specific meteorological requirements. It provides an open‑source alternative reference for building domestically operated high‑resolution AI‑precipitation forecasting systems.

\end{document}